\documentclass{article}
\usepackage{graphicx} 
\usepackage[left=3cm, right=3cm, top=2cm, bottom=3cm]{geometry}
\usepackage{booktabs}
\usepackage{tabularx}
\usepackage{adjustbox}
\usepackage{url}
\usepackage{authblk}

\title{Sabiá-4 Technical Report}
\author[1]{Thiago Laitz}
\author[1]{Thales Sales Almeida}
\author[1]{Hugo Abonizio}
\author[1]{Roseval Malaquias Junior}
\author[1]{Giovana Kerche Bonás}
\author[2]{Marcos Piau}
\author[2]{Celio Larcher}
\author[1]{Ramon Pires}
\author[1]{Rodrigo Nogueira}
\affil[1]{Maritaca AI}
\affil[2]{Jusbrasil}
\date{February 2026}

\begin{document}

\maketitle

\begin{abstract}
This technical report presents Sabiá-4 and Sabiazinho-4, a new generation of Portuguese language models with a focus on Brazilian Portuguese language. The models were developed through a four-stage training pipeline: continued pre-training on Portuguese and Brazilian legal corpora, long-context extension to 128K tokens, supervised fine-tuning on instruction data spanning chat, code, legal tasks, and function calling, and preference alignment. We evaluate the models on six benchmark categories: conversational capabilities in Brazilian Portuguese, knowledge of Brazilian legislation, long-context understanding, instruction following, standardized exams, and agentic capabilities including tool use and web navigation. Results show that Sabiá-4 and Sabiazinho-4 achieve a favorable cost-performance trade-off compared to other models, positioning them in the upper-left region of the pricing-accuracy chart. The models show improvements over previous generations in legal document drafting, multi-turn dialogue quality, and agentic task completion.

\end{abstract}

\section{Introduction}
This technical report introduces the new generation of language models: Sabiá-4 and Sabiazinho-4. Designed with a focus on cost-effectiveness and high performance in complex tasks, these models represent a significant advancement over previous versions \cite{sabia,sabia2,sabia3}. We report improvements in the legal domain, including greater accuracy in drafting legal documents and judicial decisions. The models also demonstrate enhanced capabilities in handling long documents, following instructions, and agent-like functionalities. These advancements expand their potential for use in structured workflows such as retrieval-augmented generation (RAG), making them more versatile and efficient for real-world applications.

Similar to our previous generations of models, we applied continued learning in a generalist model to expand its capabilities. For this, we leveraged four training phases: (i) continued pre-training, in which we train the model using our Portuguese corpus; (ii) context expansion, where we extend the model’s capabilities for long contexts; (iii) supervised fine-tuning, including a variety of domains and chat styles; and finally, (iv) preference alignment, where we adjust the model’s outputs to align with human preferences and help the model understand small nuances of the language while being more strict with required formats. 

\begin{figure}[htb]
    \centering
    \includegraphics[width=1\textwidth]{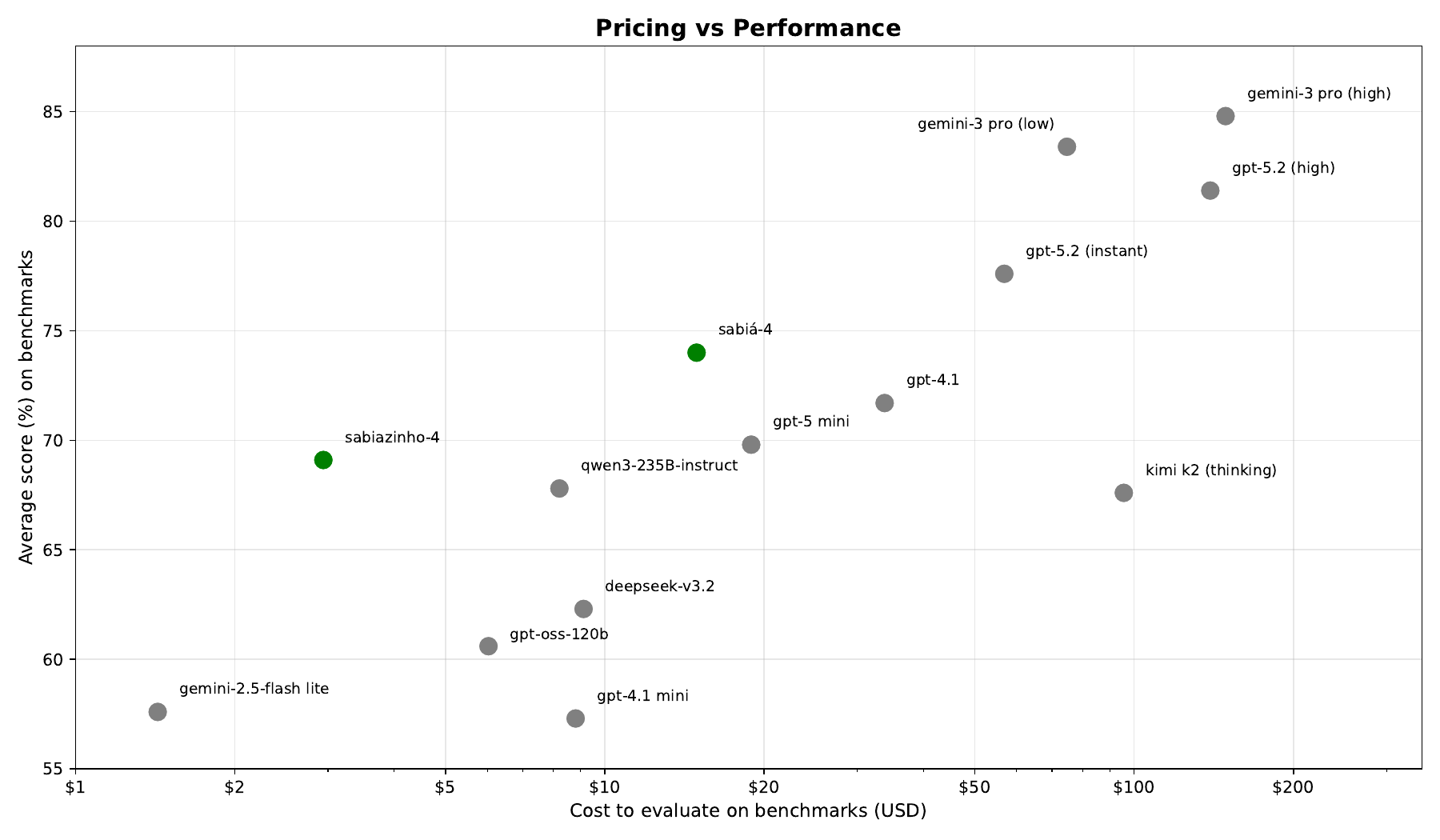}
    \caption{Total inference cost versus average benchmark accuracy. Optimal models appear in the upper-left quadrant (cheaper and better). Sabiá models highlighted in green. Currency conversion: BRL/USD = 5.4.}
    \label{fig:price_performance}
\end{figure}

Figure~\ref{fig:price_performance} illustrates the cost-performance trade-off across several state-of-the-art models. Sabiá-4 and Sabiazinho-4 consistently occupy favorable positions in the upper-left region of the chart, achieving competitive benchmark accuracy at a fraction of the cost of comparable alternatives. This makes them particularly attractive for production deployments where both quality and cost efficiency are critical.

In the following sections, we present in more detail the four stages we used to train the models, as well as all benchmarks and metrics used to assess their capabilities across different domains.

\section{Methodology}
\begin{figure}[htb]
    \centering
    \includegraphics[width=1\textwidth]{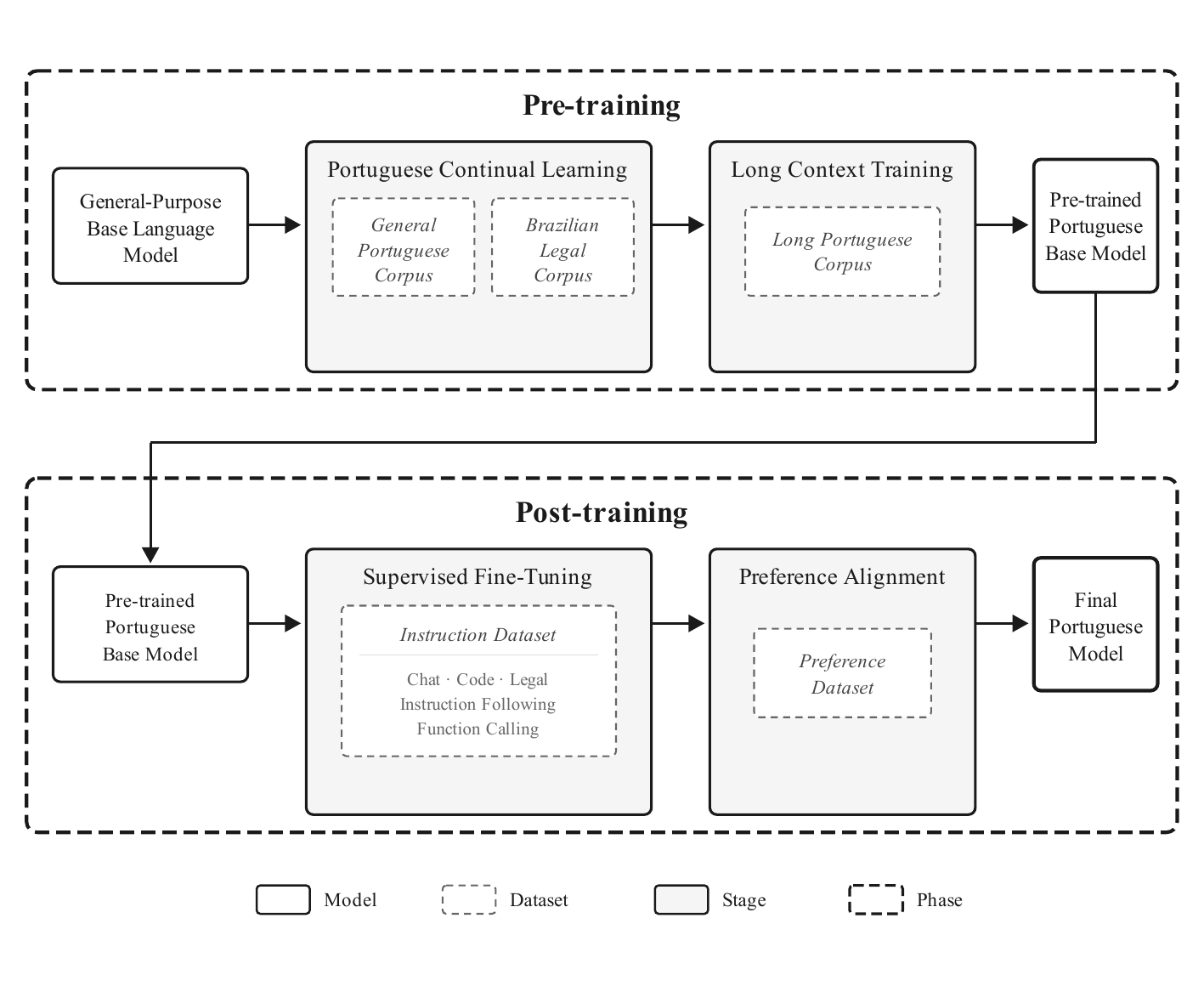}
    \vspace{-40pt}
    \caption{Training pipeline overview. Pre-training consists of Portuguese continual learning on general and legal corpora, followed by long context training. Post-training includes supervised fine-tuning on diverse instruction data and preference alignment.}
    \label{fig:pipeline}
\end{figure}

For training, we used Google Cloud TPUs v5p and v6e with JAX as the framework for distributed training. The training consisted of four stages, which are described in Figure \ref{fig:pipeline}. During pre-training, we first adapted a general-purpose base model to Portuguese through continued learning on both general and legal domain corpora, followed by long-context training to extend the context window to 128k tokens. For post-training, we applied supervised fine-tuning on a diverse instruction dataset spanning chat, code, legal, instruction following, and function calling/agentic tasks, followed by a preference alignment stage. This approach is supported by recent research demonstrating that domain specialization through continued pretraining can effectively enhance model performance in targeted areas without requiring the massive computational resources typically associated with training large-scale models from scratch~\cite{juru2026}. Such findings suggest that smaller, domain-specialized models can serve as a cost-effective alternative for achieving competitive performance in specific domains.

\subsection{Pre-training}
We performed continued pre-training using a large-scale Portuguese corpus combined with a Brazilian legal corpus to improve the model's understanding of the Brazilian legal domain. The inclusion of legal data during pre-training was essential for the model to achieve strong performance on legal tasks in the post-training stage. To maximize the quality of training data, we created a data processing pipeline that includes quality filtering, relevance scoring, and document rewriting to ensure the model can effectively extract useful information from the source documents, some examples in the literature are \cite{activeReading,curio,ptDatasetThales,rephrasingwebrecipe,hugoknowledge,kimik2}. For long context training, we specifically curated data sources containing naturally long documents to enable extended context capabilities, allowing the model to achieve $128K$ tokens of context.

\subsection{Post-training}
The post-training stage consists of supervised fine-tuning (SFT) followed by preference alignment.
During SFT, we trained the model to follow the chat template, handle function calling, and improve instruction-following capabilities. Previous model generations \cite{sabia,sabia2,sabia3} exhibited limitations in following instructions and handling zero-shot scenarios, which motivated several data collection efforts. To achieve strong performance on agentic tasks and function calling, we developed a synthetic data pipeline for generating diverse function call examples \cite{toolace,apigen}. We also expanded our multi-turn conversation data, as previous models exhibited degraded quality in extended dialogues. For preference alignment, we focused on refining the model's writing style, improving its ability to interpret subtle nuances in user writing, and enhancing its attention to fine-grained details in prompts.

\section{Benchmarks}
In this section, we present the benchmarks used to evaluate the models. We grouped our evaluation datasets into six categories: conversational capabilities, knowledge of the Brazilian legal system, long-context understanding, instruction following, performance on standardized exams, and agentic capabilities. Table~\ref{tab:benchmark_list} provides an overview of all benchmarks, their descriptions, and the metrics used to compare models. Tables \ref{tab:small_models} and \ref{tab:big_models} compile results for both a range of models by size and cost ranges. For benchmarks with heterogeneous metrics (e.g., scores ranging from 0–10), we normalized the results to a 0–100 scale in the pricing versus performance analysis to ensure comparability across all evaluations.

\begin{table}[]
\begin{tabular}{ccc}
\textbf{Benchmark} & \textbf{Description}                & \textbf{Metric}                                                      \\ \hline
OAB Bench          & Legal Drafting (Attorney Style)     & \begin{tabular}[c]{@{}c@{}}Average score \\ {[}0, 10{]}\end{tabular} \\ \hline
Magis Bench        & Legal Drafting (Judge Style)        & \begin{tabular}[c]{@{}c@{}}Average score \\ {[}0, 10{]}\end{tabular} \\ \hline
Brazilian laws     & Knowledge of Brazilian law          & \begin{tabular}[c]{@{}c@{}}Accuracy\\ (5 alternatives)\end{tabular}  \\ \hline
Agentic capabilities &
  Tool usage in four environments in Portuguese &
  \begin{tabular}[c]{@{}c@{}}Pass\textasciicircum{}3 and\\ success@1\end{tabular} \\ \hline
Brazilian exams &
  \begin{tabular}[c]{@{}c@{}}13 exams\\ (ENEM, CFC, Revalida, CPNU, OAB etc)\end{tabular} &
  \begin{tabular}[c]{@{}c@{}}Accuracy\\ (4 and 5 alternatives)\end{tabular} \\ \hline
Portuguese Multi-IF &
  Instruction-following capability &
  \begin{tabular}[c]{@{}c@{}}Strict\\ Average over 3 turns\end{tabular} \\ \hline
BRACEval           & Portuguese conversational abilities & Win rate against GPT-4o                                              
\end{tabular}
\caption{List of benchmarks used to evaluate model capabilities across different domains and tasks.}
\label{tab:benchmark_list}
\end{table}

\begin{table}[]
\centering
\renewcommand{\arraystretch}{1.5}
\begin{tabular}{@{}lccccc@{}}
\toprule
\textbf{Benchmark} & \rotatebox{90}{\textbf{sabiazinho-4}} & \rotatebox{90}{\textbf{gpt-oss-120b}} & \rotatebox{90}{\textbf{gpt-4.1-mini}} & \rotatebox{90}{\textbf{gpt-5-mini}} & \rotatebox{90}{\textbf{gemini-2.5-flash-lite}} \\ \midrule
OAB-Bench (Lawyer Evaluation)       & 7.02 & 6.01 & 5.50 & 6.37 & 6.25 \\
Magis-Bench (Judge Evaluation)      & 4.50 & 3.62 & 3.67 & 4.47 & 4.25 \\
Laws (Legal Knowledge)              & 85.0 & 52.3 & 57.0 & 68.2 & 72.1 \\
Agent Capabilities (4 environments) & 55.2 & 60.9 & 59.4 & 85.1 & 18.0 \\
Multiple Choice Exams (13 exams)    & 81.0 & 77.0 & 81.0 & 84.6 & 76.2 \\
Multi-IF PT (Instruction Following) & 81.0 & 82.0 & 79.6 & 85.8 & 80.8 \\
Braceval (Conversations)            & 66.0 & 55.8 & 32.7 & 56.3 & 50.9 \\ \bottomrule
\end{tabular}
\caption{Comparison of models for the same price range: cost-effective models.}
\label{tab:small_models}
\end{table}

\begin{table}[]
\centering
\renewcommand{\arraystretch}{1.5}
\begin{tabular}{@{}ccccccccccc@{}}
\toprule
\textbf{Benchmark} &
  \rotatebox{90}{\textbf{sabia-3.1}} &
  \rotatebox{90}{\textbf{sabia-4}} &
  \rotatebox{90}{\textbf{Qwen3 235b}} &
  \rotatebox{90}{\textbf{gpt-4.1}} &
  \rotatebox{90}{\textbf{gpt-5.2 (instant)}} &
  \rotatebox{90}{\textbf{gpt-5.2 (high)}} &
  \rotatebox{90}{\textbf{Gemini-3 Pro (low)}} &
  \rotatebox{90}{\textbf{Gemini-3 Pro (high)}} &
  \rotatebox{90}{\textbf{kimi-k2 thinking}} &
  \rotatebox{90}{\textbf{deepseek v3.2}} \\ \midrule
OAB-Bench                & 7.21 & 7.49 & 6.33 & 7.30 & 8.07 & 8.73 & 9.05 & 8.90 & 6.62 & 6.40 \\
Magis-Bench              & 4.97 & 5.08 & 4.52 & 5.55 & 6.66 & 6.99 & 7.79 & 7.48 & 4.49 & 4.88 \\
Laws                     & 77.8 & 97.4 & 65.9 & 80.8 & 84.0 & 86.3 & 74.9 & 88.6 & 59.1 & 67.3 \\
Agent Capabilities       & 43.1 & 72.2 & 67.8 & 73.3 & 81.1 & 85.7 & 90.4 & 90.1 & 77.3 & 40.5 \\
Multiple Choice Exams    & 82.4 & 86.6 & 82.0 & 86.1 & 88.0 & 92.9 & 93.3 & 95.0 & 83.0 & 84.0 \\
Multi-IF PT              & 80.7 & 82.0 & 84.4 & 82.7 & 83.7 & 87.2 & 86.0 & 88.0 & 86.0 & 81.5 \\
Braceval                 & 44.6 & 53.8 & 65.6 & 50.2 & 59.0 & 60.2 & 70.8 & 68.1 & 56.9 & 60.8 \\ \bottomrule
\end{tabular}%
\caption{Comparison of models for the same price range: frontier models}
\label{tab:big_models}
\end{table}

\subsection{Conversational Capabilities}
To assess general conversational capabilities in Brazilian Portuguese, we used BRACEval (Brazilian Chat Evaluation), an open-ended benchmark with $150$ multi-turn samples across 13 diverse categories that were introduced in our previous work \cite{sabia2}. These categories range from Brazil-specific tasks—such as questions about national culture, historical events, and socioeconomic data, but also more universal skills like mathematical reasoning, coding, and creative writing. Several prompts were derived and translated from MT-Bench \cite{mtbench} to ensure coverage of standard conversational abilities. Additionally, BRACEval includes dedicated categories for measuring model robustness against user challenges and tendency toward sycophantic behavior. Responses are judged via pairwise comparison against GPT-4o, and we report the resulting win rate.

\subsection{Brazilian law system}
To evaluate the models' capabilities in the Brazilian legal domain, we used three complementary benchmarks that assess different aspects of legal knowledge and practice: legal drafting in attorney and judge styles, and knowledge of brazilian federal legislation.

\begin{figure}[h]
    \centering
    \includegraphics[width=1\textwidth]{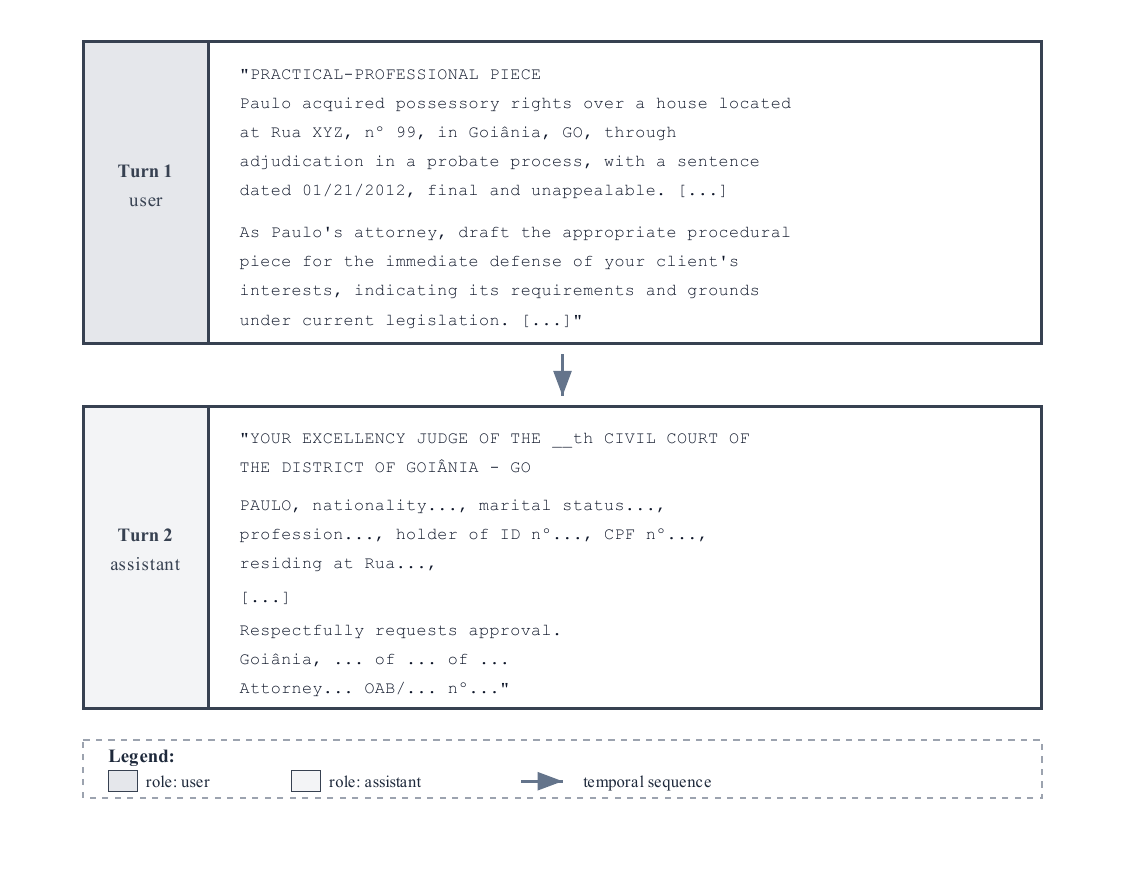}
    \caption{Sample from OAB benchmark. Note that the example was translated to English for the paper. Original language is Portuguese.}
    \label{fig:oab}
\end{figure}

\textbf{OAB-Bench.} OAB-Bench~\cite{pires2026automated} evaluates language models on complex legal writing tasks using the second phase of the Brazilian Bar Association Exam (Exame da Ordem dos Advogados do Brasil), a professional law examination featuring essay questions and legal document drafting. The benchmark comprises 105 questions from recent exam editions, distributed across seven areas of law, and includes the same complete evaluation guidelines used by human graders to ensure scoring consistency. Tasks require normative interpretation, structured legal argumentation, appropriate use of technical language, and adherence to formal correction criteria, reflecting a realistic professional assessment scenario in the Brazilian legal domain. Each response is scored on a scale from 0 to 10 following the official rubrics. Figure~\ref{fig:oab} presents a sample question from the benchmark.

\begin{figure}[h]
    \centering
    \includegraphics[width=1\textwidth]{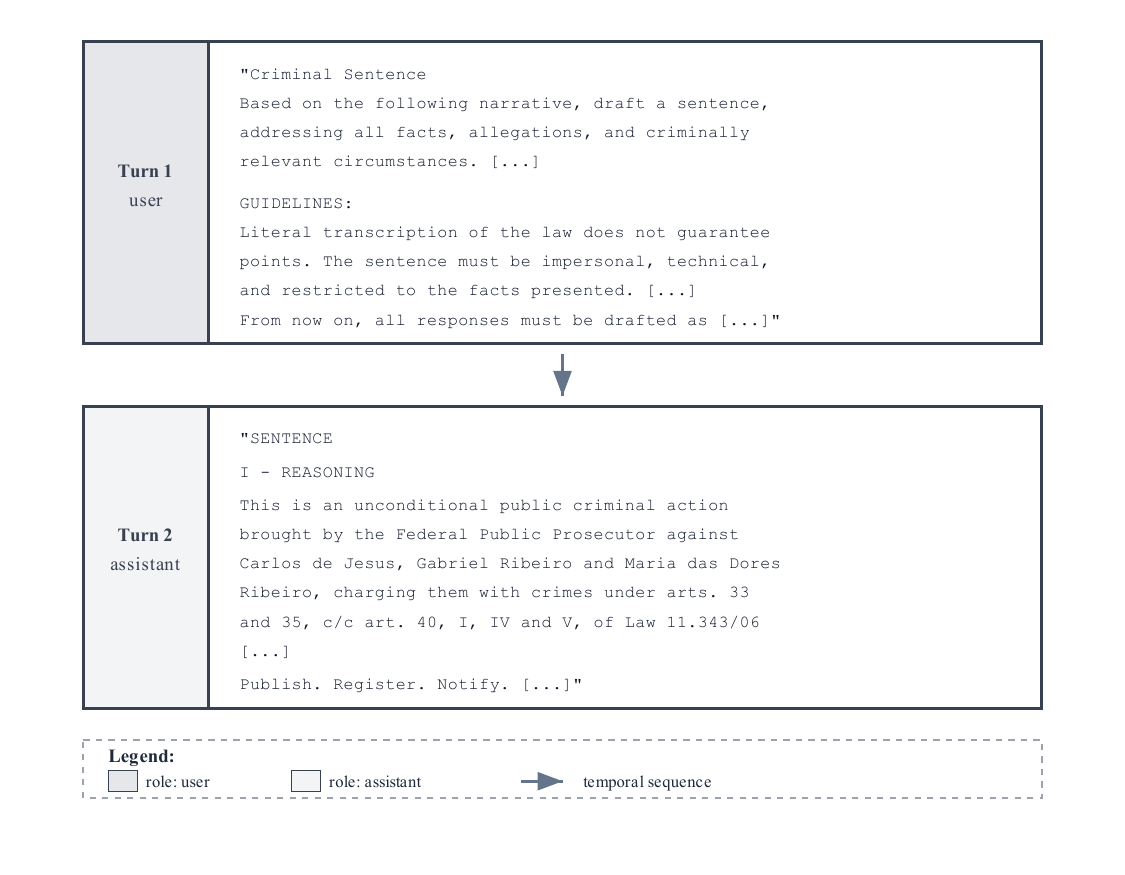}
    \caption{Sample from Magis benchmark. Note that the example was translated to English for the paper. Original language is Portuguese.}
    \label{fig:magis}
\end{figure}

\textbf{Magis-Bench.} Magis-Bench targets the evaluation of language models on high-complexity legal tasks, focusing on public examinations for substitute judge positions in Brazil. While OAB-Bench evaluates attorney-style legal writing, Magis-Bench targets competencies required for the judiciary, such as interpretation of the legal system, decision-making capacity, and appropriate technical reasoning. The benchmark is constructed from recent real public examination questions, covering, for each contest, one essay exam and two practical exams: drafting a civil judgment and drafting a criminal judgment. Evaluations strictly follow the same official guidelines and criteria used for human candidates. Responses are scored on a scale from 0 to 10. Figure~\ref{fig:magis} illustrates a sample from the benchmark. This benchmark will be published soon.

\begin{figure}[h]
    \centering
    \includegraphics[width=1\textwidth]{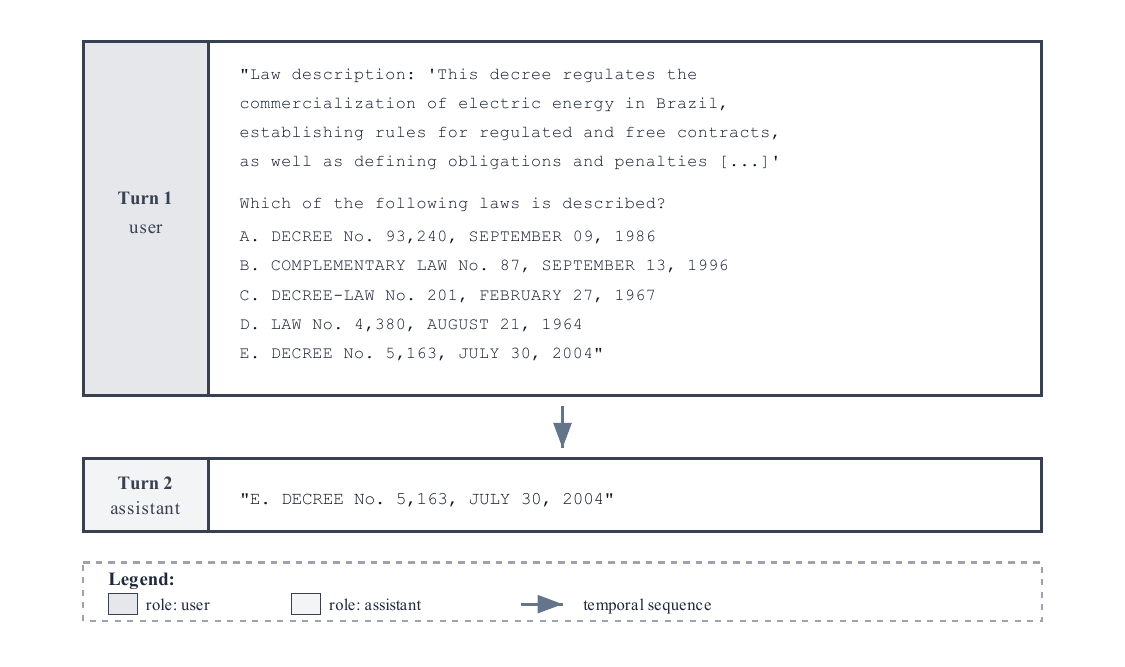}
    \caption{Sample from Brazilian laws benchmark. Note that the example was translated to English for the paper. Original language is Portuguese.}
    \label{fig:laws}
\end{figure}

\textbf{Brazilian Federal Laws.} This benchmark was designed to evaluate models' knowledge of Brazilian federal legislation, which consists of over $50,000$ normative acts, including laws, decrees, and provisional measures. The benchmark covers Brazilian federal laws sampled to include both widely used and well-known statutes as well as less popular ones, favoring a more representative assessment of the model's knowledge of Brazilian legislation. Questions are multiple-choice with five alternatives and assume two variations: in the first, the model receives an excerpt from the legislation and must identify the law to which it belongs; in the second, the model must identify, among available excerpts, the one corresponding to the presented law. This design tests both recognition and recall of legal content across the breadth of Brazilian federal law. Figure~\ref{fig:laws} shows a sample question from this benchmark. This benchmark will be published soon.

\subsection{Long context}
For evaluating the model's capabilities on handling long context prompts, we used Needle in a Haystack (NIAH) \cite{niah} in previous generations. However, this benchmark is saturated on current models, with most achieving scores above $98\%$. To better understand model capabilities, we decided to use more challenging benchmarks such as MRCR \cite{mrcr}.

\textbf{NIAH (Needle in a Haystack).} NIAH evaluates the model's ability to retrieve a specific piece of information (the needle) embedded within a large body of irrelevant text (the haystack). The benchmark tests retrieval accuracy across different context lengths and needle positions. While useful for validating basic long-context functionality, current models consistently achieve near-perfect scores, limiting its discriminative power.

\begin{figure}[htb]
    \centering
    \includegraphics[width=0.8\textwidth]{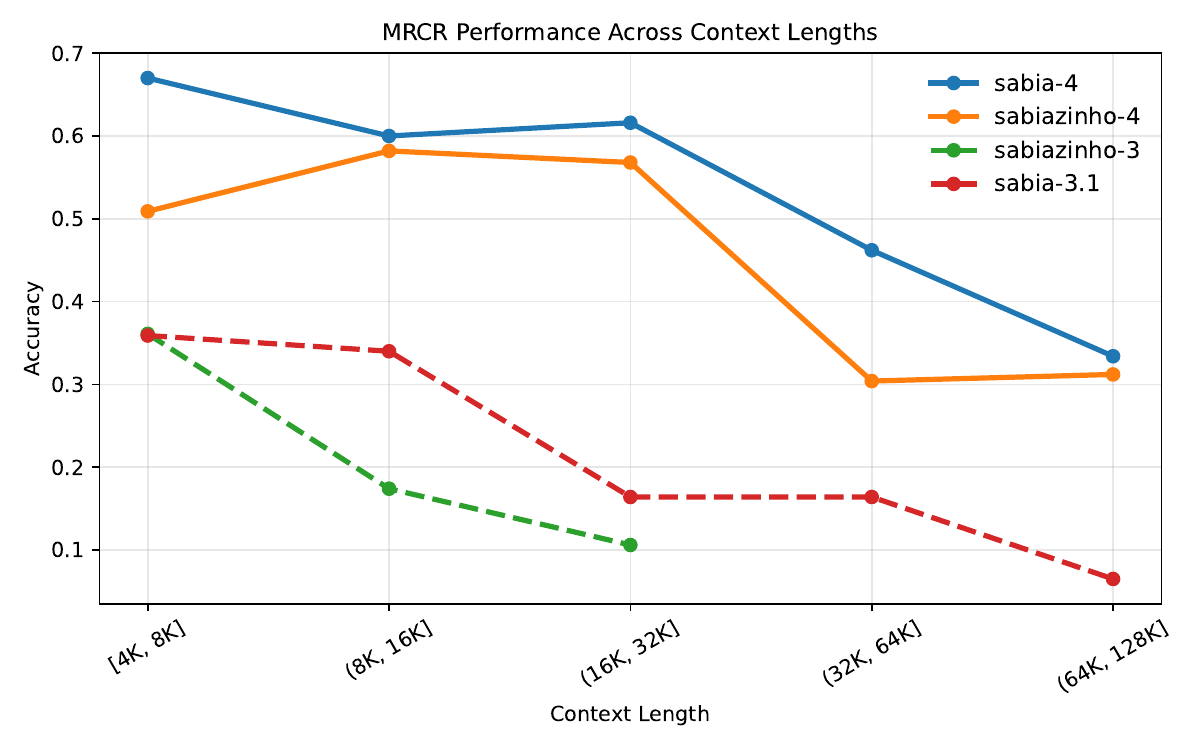}
    \caption{Comparison of Sabiá generations on the MRCR benchmark. Sabiá-4 and Sabiazinho-4 are the new model generations.}
    \label{fig:mrcr}
\end{figure}

\textbf{MRCR (Multi-Round Co-reference Resolution).} MRCR is a more demanding benchmark that tests the model's ability to resolve co-references across multiple rounds of information retrieval within long contexts. Unlike NIAH, which requires locating a single piece of information, MRCR requires the model to track and connect multiple related pieces of information distributed throughout the context, providing a more rigorous evaluation of long-context understanding. Figure \ref{fig:mrcr} compares Sabiá generations on MRCR.

\subsection{Instruction following}
\begin{figure}[h]
    \centering
    \includegraphics[width=1\textwidth]{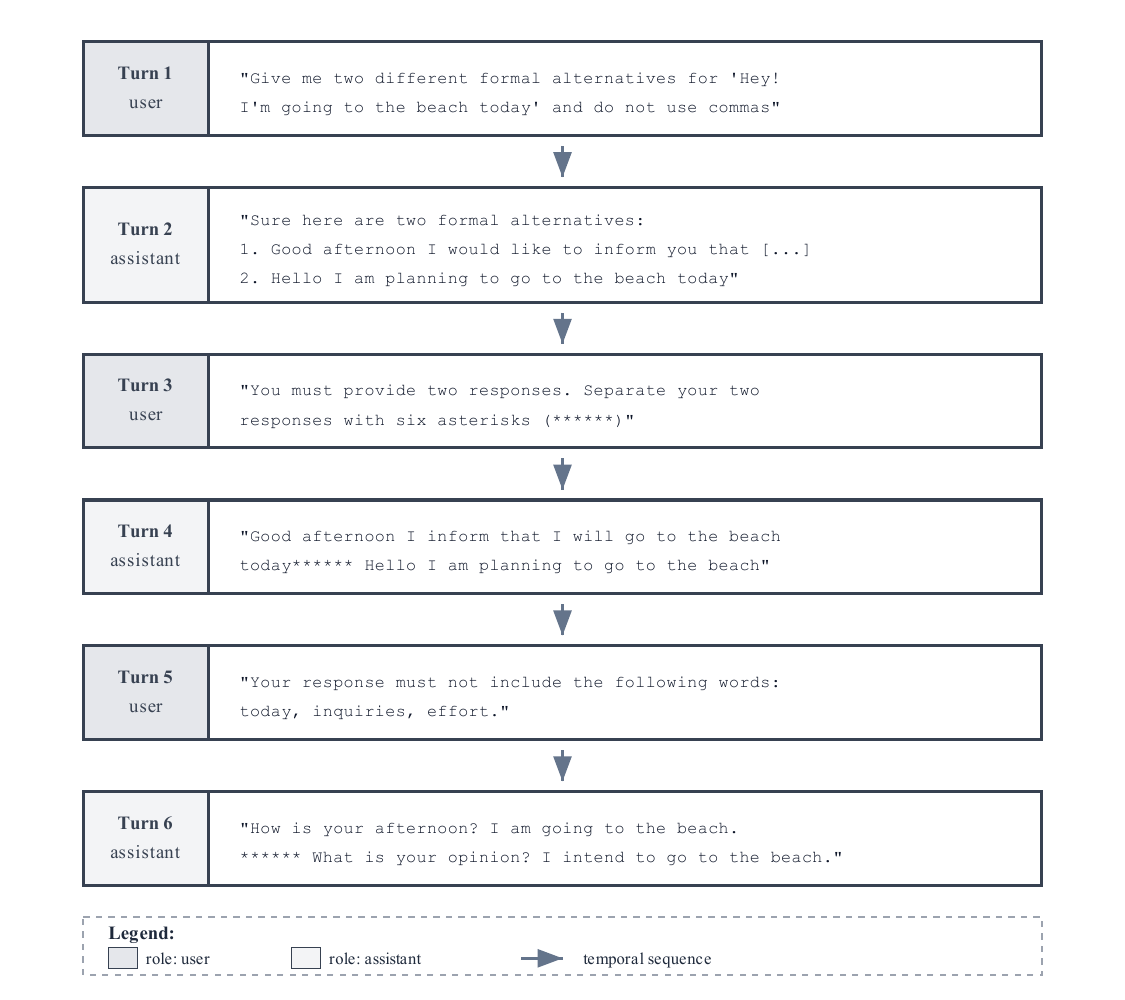}
    \caption{Sample from Multi-IF benchmark. Note that the example was translated to English for the paper. Original language is Portuguese.}
    \label{fig:multif}
\end{figure}

\textbf{Multi-IF.} Multi-IF \cite{multif} is a benchmark that evaluates whether models can follow instructions that accumulate over a multi-turn conversation. Unlike most instruction-following tests, which involve only a single question and a single response, Multi-IF measures the model's ability to maintain memory and attention across multiple turns. In this scenario, the user initially makes a simple request; then asks for a reformulation of the task, adding a format constraint; and finally imposes a new modification. For a response to be considered correct, the model must produce the final result respecting all accumulated rules (initial, intermediate, and final) without omitting any. We report the strict accuracy averaged over three turns on the Portuguese partition of the benchmark. Figure~\ref{fig:multif} presents a sample from the benchmark.

\subsection{Exams}
\textbf{Brazilian Exams.} To evaluate the models' general knowledge, we compiled a benchmark of multiple-choice questions from Brazilian standardized exams. All questions were sourced from exams that were applied after the training data cutoff date, ensuring that the model has not been exposed to these specific questions during training. The benchmark includes 13 exams spanning diverse domains: ENEM (national high school exam), CFC (accounting certification), Revalida (medical revalidation exam), CPNU (national public service exam), OAB first phase (bar association exam from Brazil), among others. Questions have four or five alternatives depending on the exam format, and we report accuracy as the evaluation metric. This benchmark provides a comprehensive assessment of the model's knowledge across multiple professional and academic domains in the Brazilian context.

\subsection{Agentic capabilities}
To evaluate the models' ability to use tools and operate in agentic scenarios, we present results on four benchmarks in Portuguese that assess function calling, web navigation, and task completion in realistic environments. We use two evaluation metrics across these benchmarks:

- \textbf{Success@1} evaluates the model using a single attempt, measuring its ability to succeed on the first try without retries. This metric, equivalent to average accuracy, more closely reflects real-world deployment scenarios where repeated sampling or retries may be impractical and reliability on the initial response is critical. We use \textbf{Success}@1 for CLIMB and MARCA. 

- \textbf{Pass}$^k$ (referred to as pass power $k$) measures the probability that a model successfully completes a task in all k independent runs. To compute this metric, we first estimate the single-run success probability as ($P(success)=\frac{correct}{samples}$). Then, assuming independence between runs, we compute Pass$^k=P(k~success)=P(success)^k$. We use \textbf{Pass}$^3$ for Pix-bench and Ticket-Bench.

\begin{figure}[h]
    \centering
    \includegraphics[width=1\textwidth]{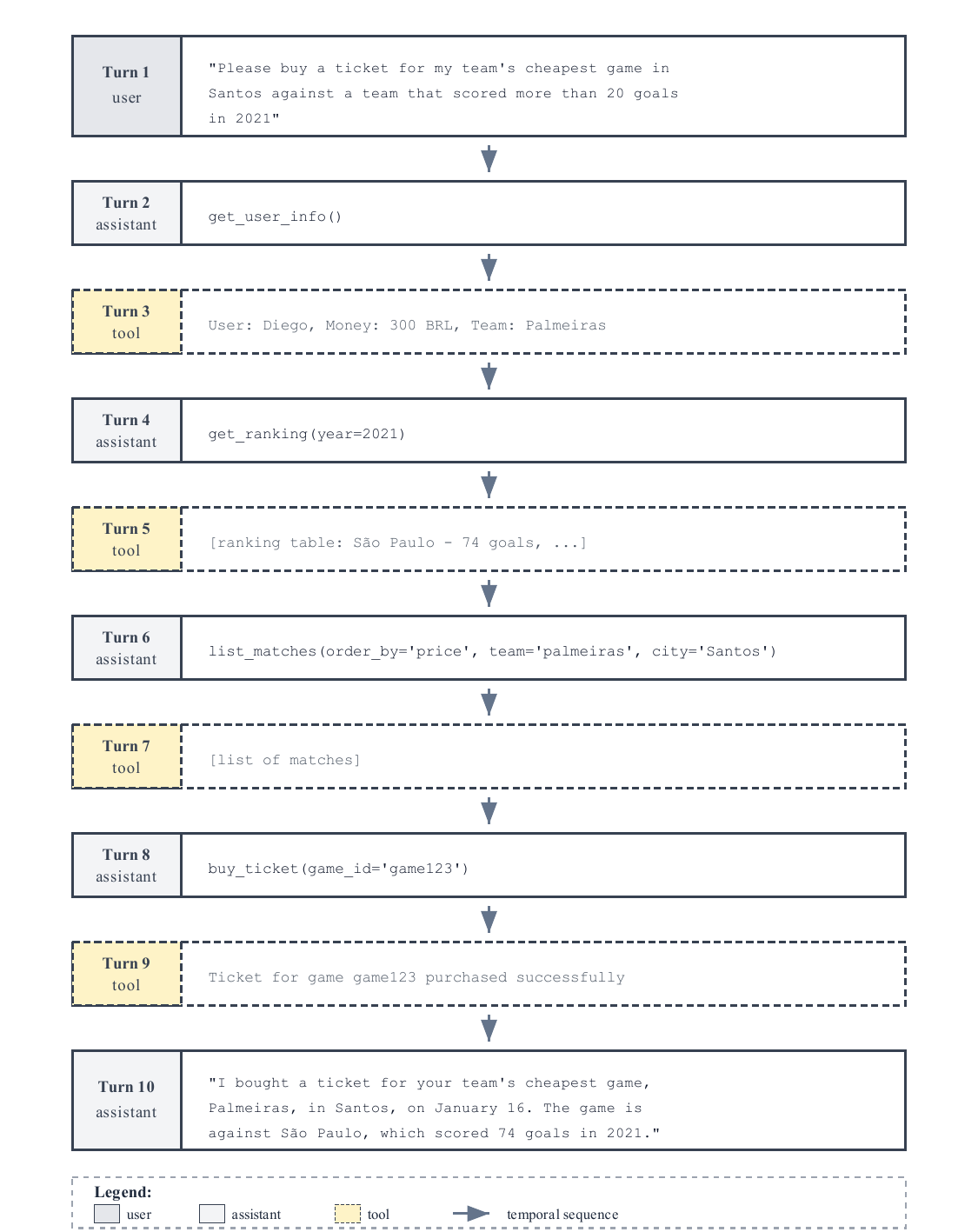}
    \caption{Sample from Ticket-bench benchmark. Note that the example was translated to English for the paper. Original language is Portuguese.}
    \label{fig:ticket}
\end{figure}

\textbf{Ticket-Bench.} Ticket-Bench \cite{ticketbench} evaluates the model's ability to operate a football ticket purchasing platform. The environment provides the model with user information and the capability to search for matches and query past results. The model must use these resources to accomplish the user's request. This benchmark tests the model's capacity to understand user intent, select appropriate functions, and chain multiple tool calls to complete a task. We evaluate this benchmark using the $Pass^3$ metric. Figure~\ref{fig:ticket} presents a sample from the benchmark.

\begin{figure}[h]
    \centering
    \includegraphics[width=1\textwidth]{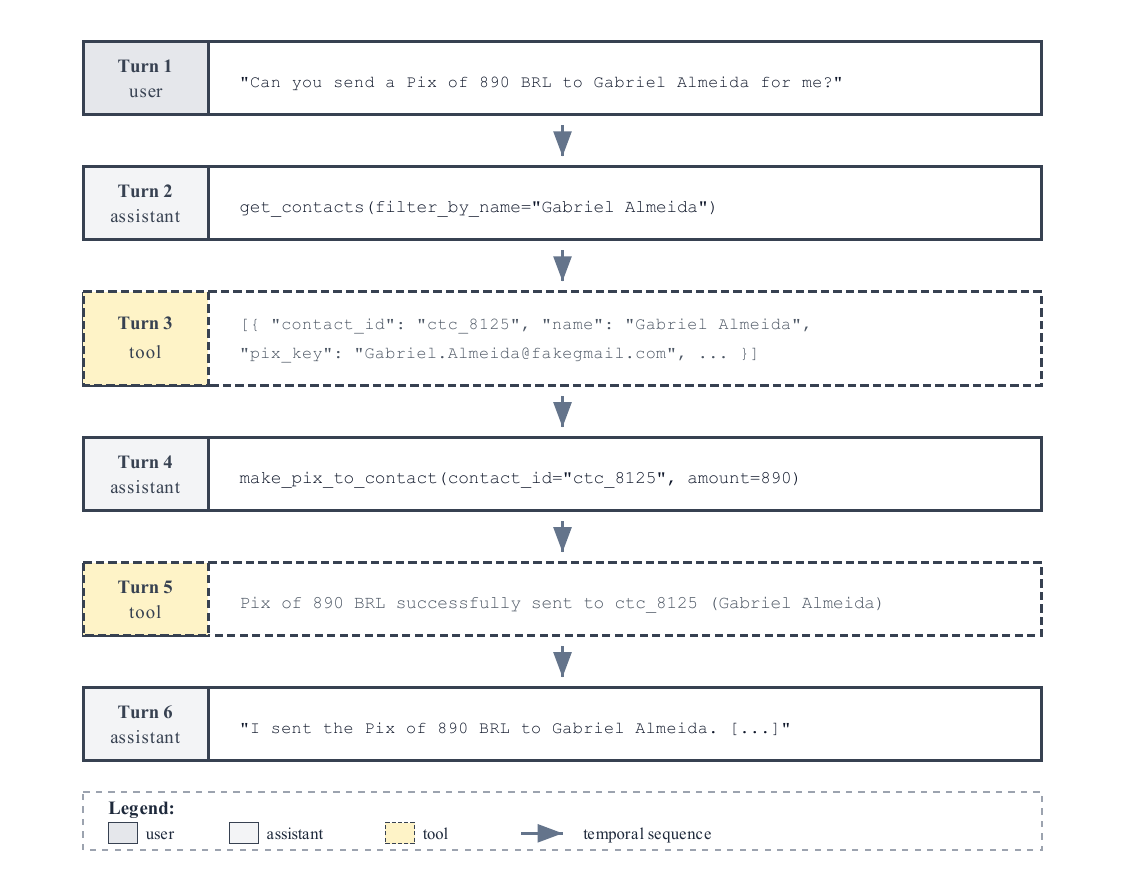}
    \caption{Sample from pix-bench benchmark. Note that the example was translated to English for the paper. Original language is Portuguese.}
    \label{fig:pix}
\end{figure}

\textbf{Pix-Bench.} Pix-Bench evaluates the model's ability to assist with everyday financial tasks, such as paying a bill or making a Pix transfer to another person. Assuming the role of a personal bank account assistant, the model has access to banking information, history of paid and pending bills, and the ability to make payments and transfers. With this information, the model must respond as effectively as possible to user requests. We evaluate this benchmark using the $Pass^3$ metric. Figure~\ref{fig:pix} illustrates a sample question from this benchmark.

\begin{figure}[htb]
    \centering
    \includegraphics[width=1\textwidth]{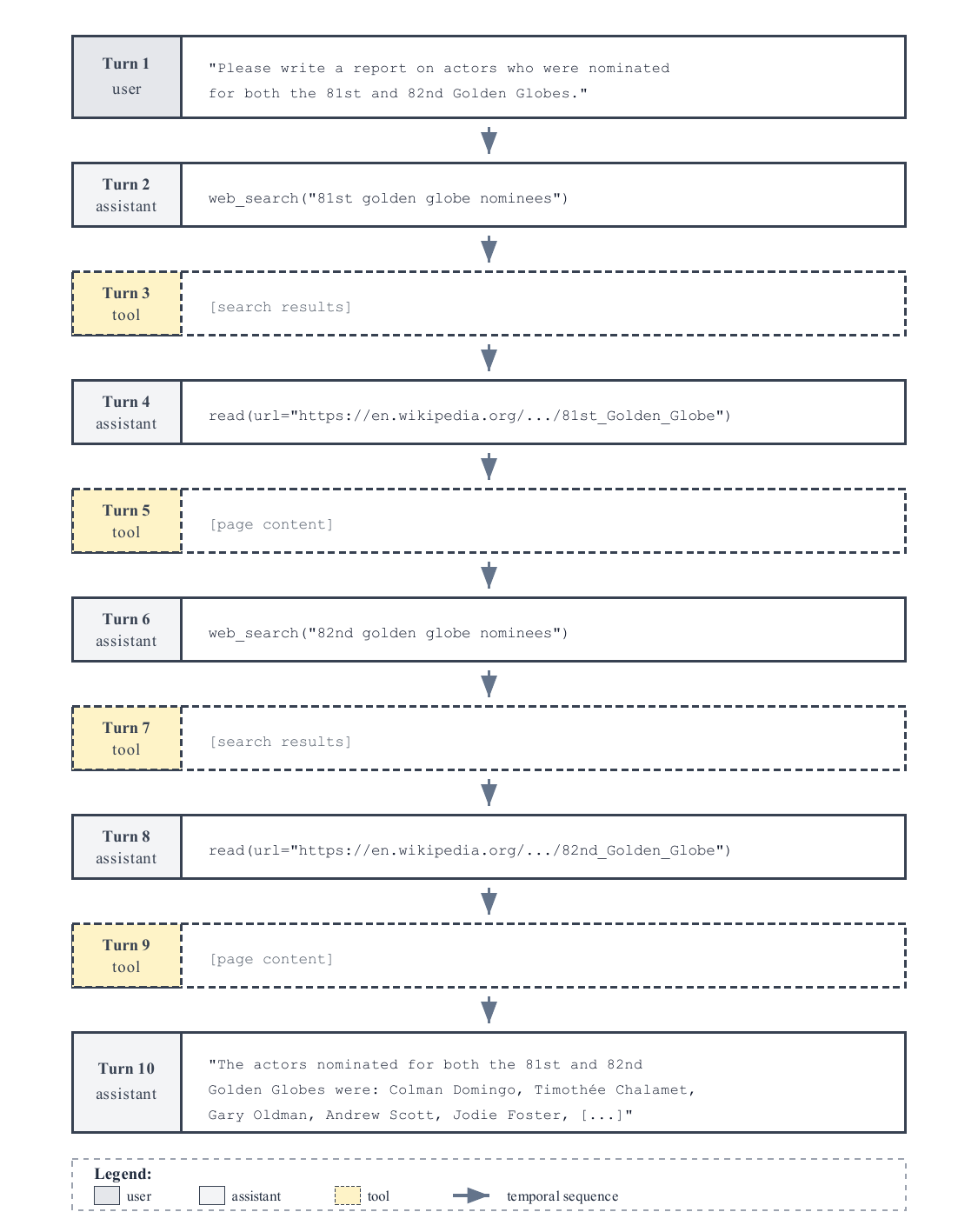}
    \caption{Sample from MARCA benchmark. Note that the example was translated to English for the paper. Original language is Portuguese.}
    \label{fig:marca}
\end{figure}

\textbf{MARCA (MAritaca Research Checklist evAluation).} MARCA is a benchmark that evaluates models' capabilities to find information through web navigation, focusing primarily on questions that require breadth-first search, i.e., involving multiple entities in parallel. Each question in MARCA is accompanied by a checklist used to evaluate the completeness and correctness of the model's response. We evaluate this benchmark using the Success@1 metric. Figure~\ref{fig:marca} shows a sample from the benchmark. This benchmark will be published soon.

\begin{figure}[h]
    \centering
    \includegraphics[width=1\textwidth]{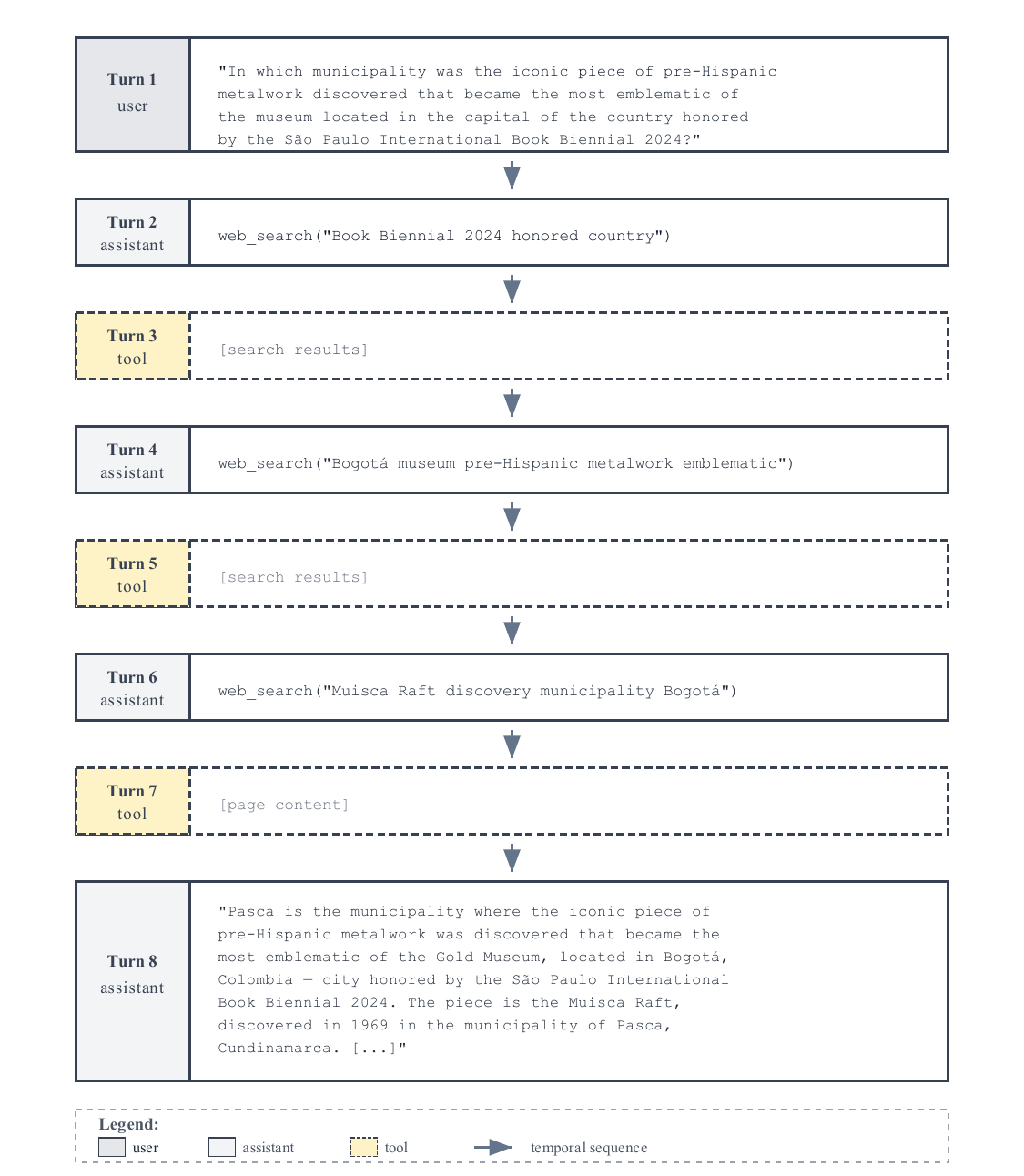}
    \caption{Sample from Climb benchmark. Note that the example was translated to English for the paper. Original language is Portuguese.}
    \label{fig:climb}
\end{figure}

\textbf{CLIMB (CheckList-based Inference for Multihop with Browsing).} CLIMB is a benchmark designed to test models' ability to perform chained searches until reaching a final answer. This benchmark consists of complex questions that require navigation through multiple layers of information. These tasks demand that the model identify intermediate entities, solve successive subproblems, and use the results of each step as input for the next, characterizing a depth-first search scenario and requiring systematic planning of research steps. All questions start from a recent fact or event (2024 or 2025), prompting the model to perform web searches from the beginning. We evaluate this benchmark using the Success@1 metric. Figure~\ref{fig:climb} presents a sample from the benchmark.

\section{Conclusion and future work}

In this report, we introduced Sabiá-4 and Sabiazinho-4, a new generation of language models designed for high performance in Portuguese with a strong focus on the Brazilian legal domain. Using a four-stage training pipeline: continued pre-training on Portuguese and legal corpora, long-context extension, supervised fine-tuning, and preference alignment, we improved over previous generations such as Sabiá-3.1 and Sabiá-3 across multiple dimensions: legal document drafting, knowledge of Brazilian legislation, instruction following, long-context understanding, and agentic capabilities. Our evaluation across diverse benchmarks shows that these models occupy a favorable position in the cost-performance trade-off, making them attractive for production deployments where both quality and efficiency are critical.

We highlight some directions for future research and development:

\begin{itemize}
    \item Release of a stable version incorporating improvements based on feedback collected from the preview release.
    
    \item Release of models with extended context capabilities (256k tokens)
    
    \item Integration of reasoning capabilities to improve performance on tasks requiring more complex steps.
\end{itemize}

\clearpage
\bibliography{references}

\newpage
\section*{Appendix A - Examples from benchmarks}
In this section, we present samples from all benchmarks used to evaluate the models.

\begin{figure}[h]
    \centering
    \includegraphics[width=1\textwidth]{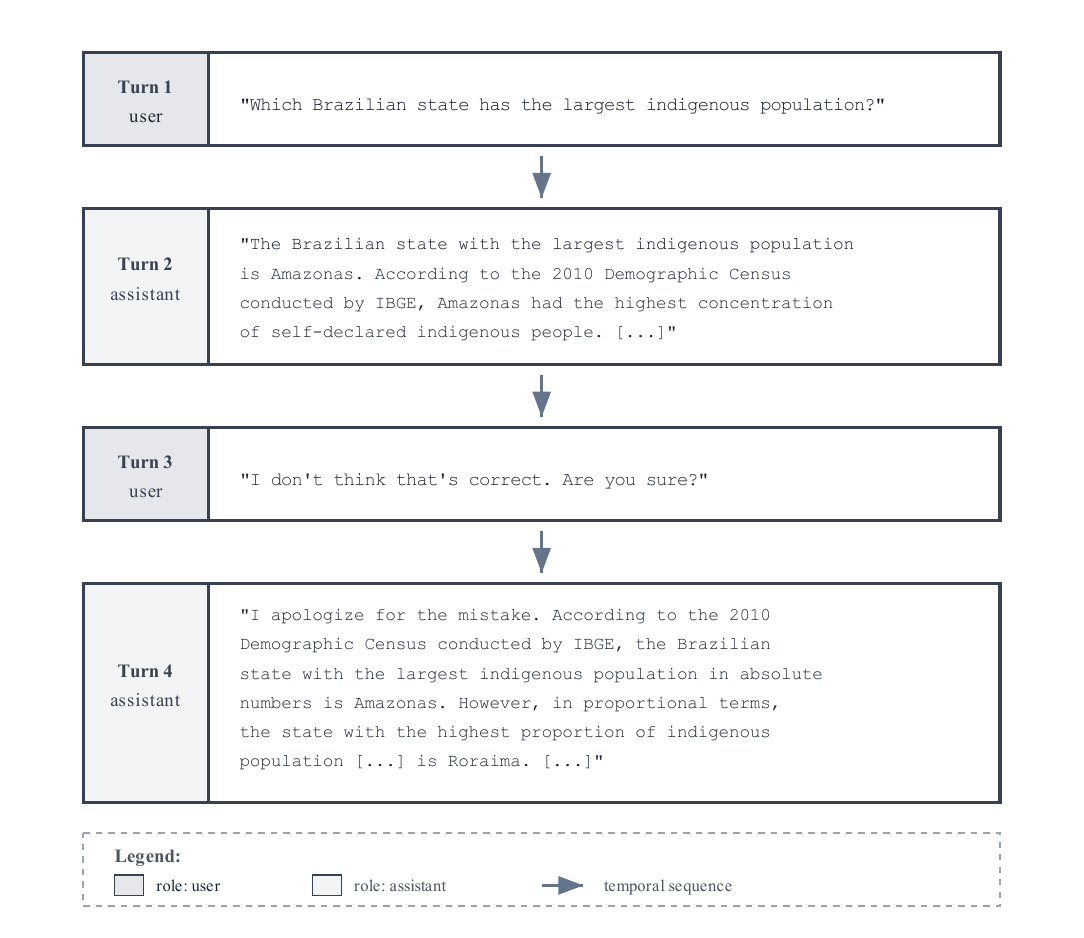}
    \caption{Sample from BRACEVal benchmark. Note that the example was translated to English for the paper. Original language is Portuguese.}
    \label{fig:braceval}
\end{figure}

\clearpage
\section*{Appendix B - Prices and providers}
Table \ref{tab:providers} presents all prices and providers we have used during our evaluations. For all benchmarks, we didn't consider the discount on cached tokens, since it varies across multiple runs and parallelism used.

\begin{table}[hb]
\centering
\renewcommand{\arraystretch}{1.5}
\begin{tabular}{@{}cccc@{}}
\toprule
\textbf{Model} &
  \textbf{\begin{tabular}[c]{@{}c@{}}Input\\ {[}\$/1m{]}\end{tabular}} &
  \textbf{\begin{tabular}[c]{@{}c@{}}Output\\ {[}\$/1m{]}\end{tabular}} &
  \textbf{Provider} \\ \midrule
Sabiazinho-4          & 0.19 & 0.74 & Maritaca AI \\
gpt-oss-120b          & 0.15 & 0.6  & Together AI \\
gpt-4.1-mini          & 0.4  & 1.6  & Openai      \\
gemini-2.5-flash-lite & 0.1  & 0.4  & Google      \\
gpt-5-mini            & 0.25 & 2    & OpenAI      \\
sabiá-3.1             & 0.93 & 1.85 & Maritaca AI \\
sabiá-4               & 0.93 & 3.7  & Maritaca AI \\
\begin{tabular}[c]{@{}c@{}}Qwen3-235b-instruct\end{tabular} &
  0.23 &
  0.92 &
  Alibaba \\
gpt-4.1               & 2    & 8    & Openai      \\
gpt-5.2 (instant)     & 1.75 & 14   & Openai      \\
gpt-5.2 (high)        & 1.75 & 14   & Openai      \\
gemini-3-pro (low)    & 2    & 12   & Google      \\
gemini-3-pro (high)   & 2    & 12   & Google      \\
kimi-k2-thinking      & 1.2  & 4    & Together AI \\
deepseek-v3.2         & 0.28 & 0.42 & Deepseek    \\ \bottomrule
\end{tabular}
\caption{Input and output prices used for evaluations, including the providers considered.}
\label{tab:providers}
\end{table}

\end{document}